%% file: m8547.tex
\documentclass{ecai} 

\usepackage{latexsym}
\usepackage{amssymb}
\usepackage{amsmath}
\usepackage{amsthm}
\usepackage{booktabs}
\usepackage{enumitem}
\usepackage{graphicx}
\usepackage{color}
\usepackage{svg}
\usepackage{eurosym}
\usepackage{caption}
\usepackage{float}
\usepackage{url}

\newcommand{\BibTeX}{B\kern-.05em{\sc i\kern-.025em b}\kern-.08em\TeX}

\begin{document}


\begin{frontmatter}


\paperid{8547} 


\title{"Where does it hurt?" - Dataset and Study on Physician Intent Trajectories in Doctor Patient Dialogues}


\author[A]{\fnms{Tom}~\snm{Röhr}}
\author[B]{\fnms{Soumyadeep}~\snm{Roy}\footnote{Equal contribution.}}
\author[C]{\fnms{Fares}~\snm{Al Mohamad}\footnotemark} 
\author[A,D]{\fnms{Jens-Michalis}~\snm{Papaioannou}} 
\author[D]{\fnms{Wolfgang}~\snm{Nejdl}}
\author[A]{\fnms{Felix}~\snm{Gers}} 
\author[A]{\fnms{Alexander}~\snm{Löser}} 
\address[A]{Berlin University of Applied Sciences, Data Science and Text-based Information Systems Group}
\address[B]{Indian Institute of Technology Kharagpur}
\address[C]{Charité – Universitätsmedizin Berlin Rheumatologie}
\address[D]{L3S Research Center, Hannover}


\begin{abstract}
In a doctor-patient dialogue, the primary objective of physicians is to diagnose patients and propose a treatment plan.
Medical doctors guide these conversations through targeted questioning to efficiently gather the information required to provide the best possible outcomes for patients.
To the best of our knowledge, this is the first work that studies physician intent trajectories in doctor-patient dialogues. We use the `Ambient Clinical Intelligence Benchmark' (Aci-bench) dataset for our study. We collaborate with medical professionals to develop a fine-grained taxonomy of physician intents based on the SOAP framework (\textbf{S}ubjective, \textbf{O}bjective, \textbf{A}ssessment, and \textbf{P}lan). We then conduct a large-scale annotation effort to label over 5000 doctor-patient turns with the help of a large number of medical experts recruited using Prolific, a popular crowd-sourcing platform.
This large labeled dataset is an important resource contribution that we use for benchmarking the state-of-the-art generative and encoder models for medical intent classification tasks.
Our findings show that our models understand the general structure of medical dialogues with high accuracy, but often fail to identify transitions between SOAP categories.
We also report for the first time common trajectories in medical dialogue structures that provide valuable insights for designing `differential diagnosis' systems. 
Finally, we extensively study the impact of intent filtering for medical dialogue summarization and observe a significant boost in performance. 
We make the codes and data, including annotation guidelines, publicly available at \url{https://github.com/DATEXIS/medical-intent-classification}.
\end{abstract}

\end{frontmatter}


\input{sections/01_introduction}

\input{sections/02_related-work}

\input{sections/03_datasets-tasks}

\input{sections/04_results}

\input{sections/05_analysis}

\input{sections/05_diag_summarization}

\input{sections/07_closing}
\graphicspath{ {images/} }



\begin{ack}
We would like to thank the reviewers for their helpful suggestions and comments. Furthermore, we would like to thank Mahmuda Akter and Keno Bressem for their support throughout this work. 

Our work is funded by the German Federal Ministry of Education and Research (BMBF) under the grant agreements 01|S23013C (More-with-Less), 01|S23015A (AI4SCM) and 16SV8857 (KIP-SDM). This work is also funded by the Deutsche Forschungsgemeinschaft (DFG, German Research Foundation) Project-ID 528483508 - FIP 12, as well as the European Union under the grant project 101079894 (COMFORT - Improving Urologic Cancer Care 
with Artificial Intelligence Solutions).
\end{ack}



\bibliography{mybibfile}

\end{document}

%% file: sections/01_introduction.tex
\section{Introduction}
\label{sec:introduction}
Doctor-patient dialogues are complex interactions where physicians must efficiently gather information, reason through differential diagnoses, and formulate treatment plans. 
While NLP research has made significant strides in tasks like medical entity recognition \citep{clinicalner}, summarization \citep{diasum}, and dialogue act classification \citep{dia_act}, most of the work in differential diagnosis modeling focuses primarily on retrospective clinical notes \citep{amega, sproto}. 
However, these notes often present a flattened and post hoc representation of patient information, neglecting the dynamic trajectories during real-time clinical conversations.
These dynamic trajectories are non-linear and an evolving process of clinical reasoning during patient encounters. 
Clinicians often revise their assessments and decisions as new information emerges throughout the consultation. 
This dynamic process involves continuous interpretation and re-interpretation of patient data, which is challenging to capture in static notes.

\begin{figure}[t]
    \centering
    \includegraphics[width=0.9\linewidth]{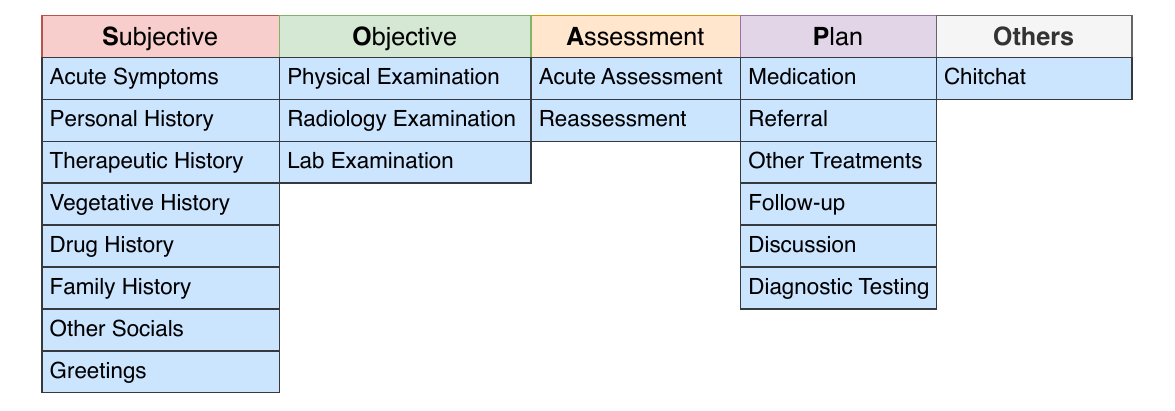}
    \caption{Proposed fine-grained physician intent taxonomy in relation to the SOAP framework, developed in consultation with medical experts. There are 8 \textit{subjective}, 3 \textit{objective}, 2 \textit{assessment}, and 6 \textit{plan} intents. We include an additional category called \textit{others}, which inherits the Chitchat intent.}
    \vspace{1em}
    \label{fig:intents}
\end{figure}

In contrast to static notes, dialogues capture the evolving intents of physicians as they navigate the complexities of a patient encounter. 
Works such as AMIE \citep{amie} demonstrate that state-of-the-art language models can effectively simulate clinical interviews by synthesizing patient interactions.
Nonetheless, how physicians transition between these steps remains largely unexplored.
To the best of our knowledge, we present the first comprehensive study of physician intent trajectories within medical dialogues using Aci-bench \citep{aci}, one of the richest datasets of doctor-patient interactions.
In close collaboration with medical professionals, we introduce a fine-grained taxonomy of physician intents as shown in Figure \ref{fig:intents}, based on the established SOAP framework \citep{soap} \textbf{as our first research contribution}.
This fine-grained taxonomy includes multiple intents per SOAP category, thus providing a highly detailed representation of how clinicians navigate patient engagements.

\begin{figure*}[t]
    \centering
    \includesvg[width=0.9\textwidth]{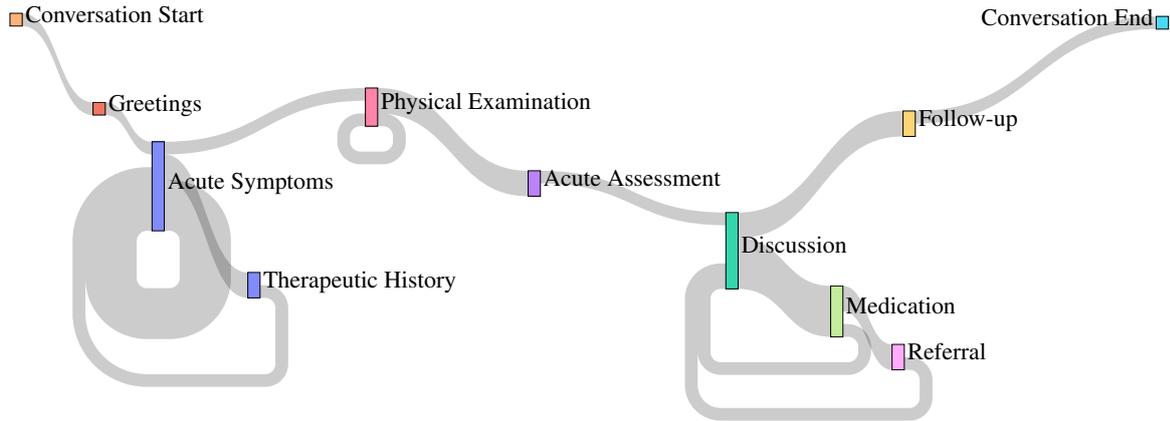}
    \caption{Physician intent trajectory during a clinical conversation. After multiple turns of subjective symptom-taking, the doctor transitions to objective examinations to conclude a clinical assessment. Finally, the conversation concludes with multiple turns for treatment planning. Please note that physicians may only use subgraphs of this general structure,  depending also on the patient's comorbidity, clinical history, and other factors.}
    \label{fig:sankey}
\end{figure*}

\textbf{As our second research contribution}, we annotate the Aci-bench dataset with the proposed intent taxonomy and make it publicly available. 
Through a large-scale crowd-sourcing effort with around \textbf{90} medical experts recruited through the Prolific platform from across the globe, we annotate more than 5,000 dialogue turns, creating a unique resource for analyzing physician trajectories in clinical conversations.
The general structure of trajectories during a differential diagnosis \citep{ddxdef} is shown in Figure \ref{fig:sankey}.

We strongly believe this annotated dataset will facilitate and encourage more research in this critical, under-explored research area. To gain a deeper understanding of intent trajectories in doctor-patient dialogues and their potential impact on the current state-of-the-art models, we perform an extensive benchmarking and characterization study, which forms our \textbf{third research contribution}. 
We evaluate generative and encoder-based models on the task of medical intent classification and next intent prediction.
Our analysis uncovers key challenges in capturing intent transitions across SOAP categories. 
Additionally, we identify common physician intent trajectories in doctor-patient dialogues.
These trajectories offer valuable insights for the design of dialogue systems to support differential diagnosis. 

As our \textbf{final research contribution}, we investigate the potential impact on current SOTA models that do not explicitly consider our proposed fine-grained physician intent taxonomy, over a critical, downstream task of dialogue-to-medical-note summarization.
We observe that filtering dialogues for physician intents improves summarization quality.
We release our dataset\footnote{\url{https://huggingface.co/DATEXIS}}, annotation guidelines, and code\footnote{\url{https://github.com/DATEXIS/medical-intent-classification}} to the community to support further research at the intersection of clinical NLP and dialogue-driven clinical decision.

%% file: sections/02_related-work.tex
\section{Related Work}
\label{sec:related}
Recent work released medical dialogue corpora to accelerate the development of medical dialogue systems (MDS).
We divide these works into two distinct groups.

\noindent \textbf{Non-annotated medical dialogues.}
 These datasets do not contain specific dialogue annotations and have either a small number of examples \citep{related_work13}, only include short dialogues \citep{related_work12}, or are not freely accessible \citep{related_work14,related_work15, related_work16, related_work17}.
Larger datasets like \citep{MedDialog} are non-English and lack real-world conversations. 

\noindent \textbf{Annotated medical dialogues.}
Works such as ReMeDi \citep{remedi}, MIE \citep{MIE}, Code-Mixed \citep{Code-mixed}, IMCS-21 \citep{oxford}, and MediTOD \citep{meditod} curate resources that align with annotations to solve MDS tasks. 
Such tasks include annotations for medical entity recognition, natural language generation, or dialogue act classification.
Our work focuses on a more detailed annotation of physician intents in dialogues guided by the SOAP taxonomy.
We use SOAP because it is a widely adopted standard for documenting clinical notes.
Therefore, our approach bridges the gap between the representation of patient interactions in dialogues and their documentation in medical notes.

\noindent \textbf{Modeling differential diagnosis in medical dialogue systems.}
Several studies, including AMIE \citep{amie}, MEDDxAgent \citep{meddxagent}, and \citet{mdagents}, investigate the application of foundation models in the differential diagnosis process. 
These works highlight that the initial dialogue phase is the most critical stage, as it shapes the quality and accuracy of subsequent diagnostic reasoning.
However, the extent to which these models effectively capture transitions in clinical reasoning remains an open research question.
Previous studies primarily focus on modeling differential diagnoses or generating physician-like dialogues. 
In contrast, our work explicitly analyzes physician intent trajectories within medical conversations.
We provide a structured framework for understanding how clinicians transition between reasoning stages in real-time interactions.

%% file: sections/03_datasets-tasks.tex
\section{Medical Intent Dataset}
\label{sec:annotation}
In this section, we present the annotation taxonomy, outline the approach used to develop the annotation guidelines, and discuss the intent annotation process.
Finally, we offer insights into the dynamics of medical dialogues across the SOAP categories.
\subsection{Dataset Construction}
\paragraph{Data pre-processing.} 
We obtain the utterances in our corpus from the role-played medical dialogue dataset Aci-bench \citep{aci}.
Aci-bench is a dialogue summarization dataset consisting of 207 dialogue-clinical note pairs. 
We select this dataset for annotation due to its comprehensive collection of doctor-patient dialogues spanning various medical specialties, with an emphasis on authentic clinical interactions.
Each dialogue is organized with clearly defined speaker roles, and we segment the dialogues into distinct doctor-patient turns according to these roles. 
We manually revise dialogues containing reversed roles or concatenated utterances to ensure accurate doctor-patient turns.
After pre-processing, we end up with \textbf{5,541} doctor-patient turns.

\paragraph{Intent taxonomy.}
\label{subsec:phases}
We design the intent classes to align with the established \textbf{S}ubjective, \textbf{O}bjective, \textbf{A}ssessment, and \textbf{P}lan (SOAP) \citep{soap} taxonomy.
Figure \ref{fig:intents} shows a comprehensive overview of all intents.
This taxonomy allows us to create intents that break down the dialogues into specific phases that are crucial for the differential diagnosis process, as we highlight in Figure \ref{fig:sankey}. 

\begin{figure*}
    \centering
    \includegraphics[width=\linewidth]{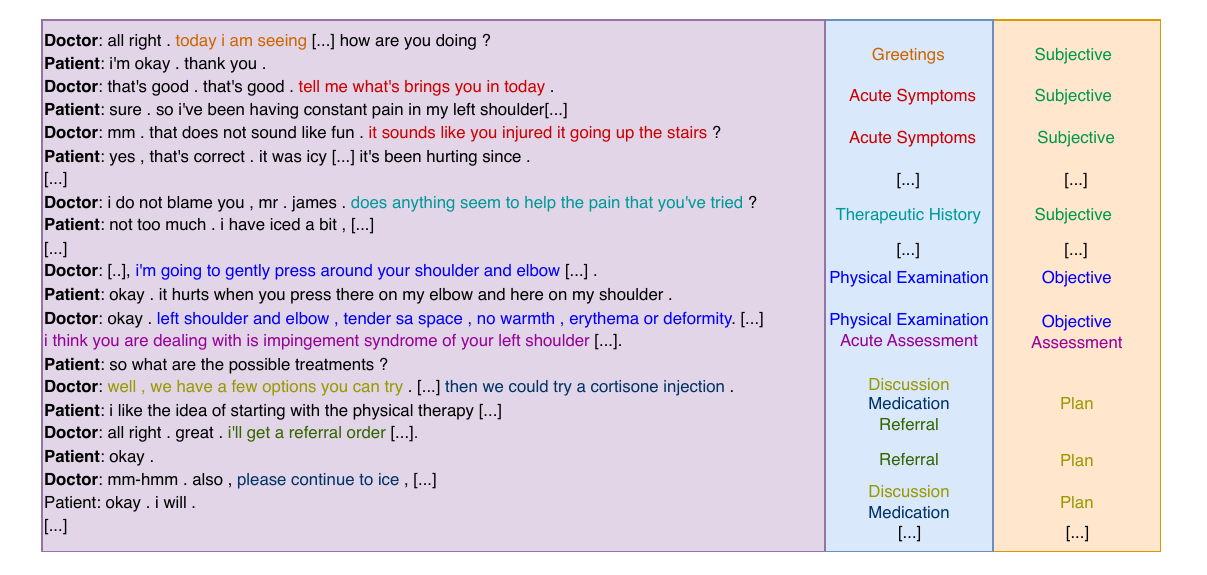}
    \caption{Excerpt of an annotated dialogue. We see that a dialogue is characterized by multiple \textit{Subjective} iterations in the beginning. The dialogue then transitions to \textit{Objective} iterations, which lead to the \textit{Assessment}. With multiple iterations in \textit{Plan}, the dialogue finishes.}
    \label{fig:diaexample}
\end{figure*}

Figure \ref{fig:diaexample} provides the excerpt of the actual dialogue shown previously in Figure \ref{fig:sankey}. Multiple, short questions by the physician at the beginning of a dialogue characterize the symptom-taking phase (\textit{Subjective}). 
We see in Figure \ref{fig:diaexample} that the physician iterates multiple times on \textit{Acute Symptoms} and asks for the \textit{Therapeutic History} of the patient.
In the examination phase (\textit{Objective}) the physician collects factual diagnostic observations, such as \textit{Physical Examination}, \textit{Radiology Examination}, and \textit{Lab Examination}.
The examination phase typically follows the symptom-taking phase and, due to its factual nature, requires less repetition than the symptom-taking phase.
The clinical assessment phase (\textit{Assessment}) involves diagnosing the patient and usually follows the examination phase.
As shown in Figures \ref{fig:sankey} and \ref{fig:diaexample}, the clinical assessment phase is precise, requiring mostly no repetitions by the physician.
Lastly, the dialogue concludes with the treatment-planning phase (\textit{Plan}), where the physician and patient discuss the proposed treatment plan.
As illustrated in Figure \ref{fig:sankey}, multiple iterations often occur during this phase. 
These loops emerge as the patient consents to or engages with the proposed plan.
This process repeats until both parties agree.

\paragraph{Annotation guidelines.} 
We collaborate with practicing physicians to develop comprehensive annotation guidelines for physician intent classification.
To ensure clarity and consistency, we iteratively refine both the intent taxonomy and the annotation guidelines. 
Each iteration involves an external annotator applying the guidelines to a small subset of the dataset, followed by a critical evaluation of ambiguities, edge cases, and potential refinements. 
The finalized guidelines contain 20 intent classes across 5 categories.
For annotation, we adhere to standard practices and initially perform the labeling in-house. 
Subsequently, we verify the accuracy of the annotations with the help of medical professionals through a crowd-sourcing platform \citep{human_annotations1, human_annotators2, human_annotators3}.
We expand on the annotation process in the supplementary material \cite{supp_material}.

\paragraph{Data verification.} 
Medical professionals, recruited through the crowd-sourcing platform Prolific\footnote{\url{https://www.prolific.com}}, verify our annotations to ensure their reliability.
Our effort achieves an annotation accuracy of 81.13\%.
We systematically review the remaining 19.87\% of cases and incorporate annotator feedback, removing samples with unresolved disagreements.
Further details on the verification process are provided in supplementary material \cite{supp_material}.

\begin{table}[!ht]
\caption{Statistics for categories per turn, category tokens per dialogue, and the most frequent intent per category in the annotated dataset. The token statistics are for doctor utterances only. A doctor spends the most turns in \textit{Subjective} symptom-taking but discusses the most in \textit{Plan}. (AS: Acute Symptoms, PE: Physical Examination, AA: Acute Assessment, D: Discussion, C: Chitchat)}
\scalebox{0.9}{
\begin{tabular}{cccccc}
\hline
                      & \textbf{Subjective} & \textbf{Objective} & \textbf{Assessment} & \textbf{Plan} & \textbf{Others} \\ \hline
\textbf{Total count}  & 2860                & 876                & 368                 & 1143          & 616             \\
\textbf{Mean count}   & 13.81               & 4.23               & 1,77                & 5.52          & 2.97            \\
\textbf{Max count}    & 36                  & 20                 & 8                   & 43            & 20              \\ \hline
\textbf{Total tokens} & 67,466              & 71,915             & 58,093              & 89,826        & 9409            \\
\textbf{Mean tokens}  & 325.92              & 347.61             & 280.64              & 433.94        & 45.45           \\
\textbf{Max tokens}   & 1045                & 1059               & 789                 & 1600          & 203             \\ \hline
\textbf{Top intent}   & AS                  & PE                 & AA                  & D             & C               \\ \hline
\end{tabular}}
\centering

\label{tab:dataset_stats}
\end{table}

\subsection{Characterization Study on Dynamics in Doctor-Patient Dialogues}
\paragraph{Doctors spend the most turns on subjective symptom-taking.} 
Table \ref{tab:dataset_stats} shows the time doctors spend per SOAP category in a dialogue with a patient.
We show that doctors invest most turns for the symptom-taking phase, with \textit{Acute Symptoms} being the most frequent intent.
In contrast, a doctor needs the least turns for the clinical assessment phase.
While the symptom-taking phase has the most turns on average, the treatment-planning phase can potentially extend over a longer period, as indicated by the maximum number of turns observed across all dialogues in Table \ref{tab:dataset_stats}.
This is mainly due to the nature of the treatment-planning phase, which often involves continuous discussions and negotiations between the doctor and patient about treatment options. 
Such interactions may require multiple iterations before both parties reach a mutual agreement.

\paragraph{Doctors speak most during treatment-planning.}
Although the average number of turns in the treatment-planning phase is lower than in the symptom-taking phase, Table \ref{tab:dataset_stats} shows that the doctor speaks the most during treatment-planning, as indicated by the mean number of tokens per category.
In this phase, \textit{Discussion} is the most frequent intent.
This underscores the difference between the one-sided process of collecting subjective symptoms and the collaborative nature of treatment planning.
During symptom collection, the doctor primarily collects information by questioning the patient. 
In contrast, treatment planning involves both the doctor and the patient actively engaging in a dynamic discussion that can evolve without a predetermined outcome.

\paragraph{Chitchat in doctor-patient dialogues is omnipresent.} 
On average, a dialogue includes more \textit{Chitchat} turns than turns in the clinical assessment phase.
However, despite their frequency, \textit{Chitchat} turns are brief and can appear in every phase of the dialogue.
These turns contain little to no informational content and can be regarded as noise, as they do not aid in the differential diagnosis process.

\paragraph{Conclusion.} Our findings on the frequency of subjective symptom-taking intents and the omnipresence of chitchat overlap with data statistics published in \citet{remedi, meditod}, and \citet{MIE}.
Similarly to our distribution, we see that the majority of entities are symptom-taking intents and that chitchat is distributed across all dialogues.
Since no related work reports utterance length statistics on the intent level, we cannot substantiate our second claim that treatment-planning utterances contain the most words on average.

%% file: sections/04_results.tex
\section{Experimental Setup}
\label{sec:experiments}
This section discusses the evaluation tasks and the baseline models used in our experiments.
Both tasks are multi-label classification tasks, and we apply stratified sampling \citep{startsampling} to produce training, validation, and test splits.
To ensure a comprehensive evaluation of our imbalanced dataset, we report both macro-AUROC and macro-Average Precision (AP). 
While macro-AUROC evaluates classification performance by measuring the area under the ROC curve, macro-AP provides a more nuanced metric by emphasizing precision and recall, particularly for underrepresented classes.
\subsection{Task Definitions}
 
\paragraph{Task: Medical intent classification.} 
The medical intent classification task assesses whether a model is capable of mapping physician utterances to medical intents. 
Each input consists of a single physician utterance, and the model is tasked with predicting one or more intents.

We show the dataset statistics for this task in Table \ref{table:int-c-statistics}.

\begin{table}[h!]
\caption{Intent classification dataset statistics after stratified splitting.}
\begin{tabular}{lcccc}
\hline
\multicolumn{1}{c}{Statistics} & All             & Train           & Val             & Test            \\ \hline
Total \# samples       & 5292     & 3886     & 646       & 760     \\ 
Avg. \# intents        & 1.41      & 1.46      & 1.27     & 1.27      \\ 
Avg. \# sections       & 1.11     & 1.32      & 1.03      & 1.04      \\ 
Avg. \# tokens per utterance  & 36.54 & 39.19  & 28.97 & 29.44  \\ \hline
\end{tabular}
\centering

\label{table:int-c-statistics}
\end{table}

\paragraph{Task: Next intent prediction.}
The next intent prediction task evaluates whether a model can predict the subsequent physician intent in the trajectory of a doctor-patient dialogue.
Each input consists of up to five preceding doctor-patient turns, and the model is tasked with predicting one or more intents associated with the next step of the physician in the sequence.
For cases where the prediction involves the first turn in the dialogue, we prepend a fixed \textit{Conversation Start} token to represent the absence of prior context.
Table \ref{table:next-i-p-statistics} presents the dataset statistics for this task.

\begin{table}[h!]
\caption{Next intent prediction dataset statistics after stratified splitting.}
\begin{tabular}{ccccc}
\hline
Statistics               & \multicolumn{1}{l}{All} & Train  & Val     & Test    \\ \hline
Total \# samples         & 5292                    & 3886   & 646     & 760     \\
Avg. \# previous intents & 5.83                    & 5.88   & 5.76    & 5.73    \\
Avg. \# previous turns   & 4.14                    & 4.16   & 4.08    & 4.08    \\
Avg. \# tokens         & 257.35                 & 258.67 & 249.74 & 257.04 \\ \hline
\end{tabular}
\centering

\label{table:next-i-p-statistics}

\end{table}

\subsection{Baseline Models}
The following encoder and decoder-only model settings apply for both tasks.

\paragraph{Encoder models.}
We select state-of-the-art clinical encoder models GatorTronS \citep{gatortron, gatortrongood} and BiomedBERT \citep{pubmedbert, copiv} and fine-tune them in two settings.
The first setting is fine-tuning on the intent classes only, whereas the second setting is a hierarchical fine-tuning.
In the hierarchical approach, the model first predicts the SOAP categories and then the intent classes.
We mask intents that do not associate to the predicted SOAP categories from the first step and calculate a loss as an average of both steps.
The optimizer is AdamW \citep{adam}.
\paragraph{Decoder-only models.}
Due to their reasonable size and state-of-the-art performance we evaluate  Llama-3.1-8B-Instruct \citep{llama}, Qwen2.5-7B-Instruct \citep{qwen}, and Phi-4-14B \citep{phi4}.
In order to adapt autoregressive models to classification tasks, we employ guided decoding and follow \citet{willard2023efficient}. 
We enforce the models to always generate an output that contains all classes paired with a boolean value that indicates the presence or absence of the class in the current utterance. 
Thus, we can replicate a discrete prediction space and apply classification metrics without the need for sophisticated post-processing of the output.

We refrain from training the decoder-only models and instead evaluate them at inference time in both zero-shot and few-shot settings.
In the zero-shot setting, we provide only a simple prompt that instructs the model to classify the current sample. 
For the few-shot setting, we additionally include \((x,y)\) examples in the prompt. 
We retrieve relevant examples by computing the BM25 \citep{bm25} score between the input \(x\) and an example corpus \(C\), where \(C\) comprises all samples from the training and validation splits.
In few-shot experiments, we incorporate the top three retrieved examples.
We provide a prompt example in the supplementary material \cite{supp_material}.

%% file: sections/05_analysis.tex
\section{Experimental Results and Discussion}
\label{sec:resultsanddiss}
We present results for all models on both tasks in Table \ref{table:benchmark_results} 
and analyze the intent-wise performance of the best-performing model. 
Furthermore, we conduct an ablation study with a fine-tuned next intent prediction model to reconstruct dialogue sequences.
\begin{table}[t]
\caption{Experimental results for all models on both tasks. We report AUROC and Average Precision (AP) macro averaged. \(\pm\) denotes the standard deviation before aggregation. Fine-tuning encoder models performs significantly better than decoder-only models.}
\begin{tabular}{lcccc}
\hline
                              & \multicolumn{2}{c|}{\textbf{Intent Classification}} & \multicolumn{2}{l}{\textbf{Next Intent Prediction}} \\ 
                              & \textbf{AUROC}  & \multicolumn{1}{c|}{\textbf{AP}}  & \textbf{AUROC}             & \textbf{AP}            \\ \hline
\multicolumn{1}{c}{\textbf{}} & \multicolumn{4}{c}{\textbf{Intent fine-tune}}                                                                    \\ 
BiomedBERT                    & 0.91\(\pm0.06\)            & \multicolumn{1}{c|}{0.63\(\pm0.21\)}         & 0.82\(\pm0.08\)                   & 0.27\(\pm0.25\)                   \\
GatortronS                    & \textbf{0.93\(\pm0.05\)}            & \multicolumn{1}{c|}{\textbf{0.69}\(\pm0.18\)}         & \textbf{0.85\(\pm0.06\)}                       & \textbf{0.37\(\pm0.25\)}                  
\\ \hline
\multicolumn{1}{c}{\textbf{}} & \multicolumn{4}{c}{\textbf{Hierarchical fine-tune}}                                                                    \\ 
BiomedBERT                    & 0.88\(\pm0.08\)            & \multicolumn{1}{c|}{0.64\(\pm0.18\)}         & \textbf{0.66\(\pm0.14\)}                   & \textbf{0.19}\(\pm0.16\)                   \\
GatortronS                    & \textbf{0.89\(\pm0.07   \)}            & \multicolumn{1}{c|}{\textbf{0.69}\(\pm0.17\)}         & 0.57\(\pm0.11\)                      & 0.10\(\pm0.10\)                   \\\hline
\multicolumn{1}{c}{\textbf{}} & \multicolumn{4}{c}{\textbf{Zero-shot}}                                                                    \\ 
Llama3.1                      & 0.56\(\pm0.07\)           & \multicolumn{1}{c|}{0.07\(\pm0.06\)}         & 0.57\(\pm0.05\)                       & 0.08\(\pm0.06\)                   \\
Phi4                          & \textbf{0.79\(\pm0.11\)}            & \multicolumn{1}{c|}{\textbf{0.28\(\pm0.14\)}}         & 0.63\(\pm0.10\)                       & 0.14\(\pm0.16\)                   \\
Qwen2.5                       & 0.73\(\pm0.11\)            & \multicolumn{1}{c|}{\textbf{0.28}\(\pm0.17\)}         & \textbf{0.66\(\pm0.10\)}                       & \textbf{0.15\(\pm0.14\)}                   \\ \hline
\multicolumn{1}{c}{\textbf{}} & \multicolumn{4}{c}{\textbf{Few-shot (3)}}                                                                \\ 
Llama3.1                      & 0.67\(\pm0.09\)            & \multicolumn{1}{c|}{0.16\(\pm0.12\)}         & 0.61\(\pm0.07\)                       & 0.12\(\pm0.10\)                 \\
Phi4                          & \textbf{0.82}\(\pm0.08\)            & \multicolumn{1}{c|}{\textbf{0.33}\(\pm0.17\)  }         & \textbf{0.65\(\pm0.09\)}                       & 0.16\(\pm0.14\)                   \\
Qwen2.5                       & 0.74\(\pm0.12\)             & \multicolumn{1}{c|}{0.32\(\pm0.23\)}         & \textbf{0.65\(\pm0.10\)}                       & \textbf{0.20\(\pm0.20\)}                   \\ \hline
\end{tabular}
\centering

\label{table:benchmark_results}
\end{table}

\subsection{Experimental Results} 
\paragraph{Intent classification.}
Fine-tuned encoder-based models consistently outperform all decoder-only models by at least
\(70.58\%\) Average Precision (AP). 
GatorTronS achieves the highest performance, closely followed by BiomedBERT with a \(9.09\%\) AP difference.
Both encoder models do not benefit from hierarchical fine-tuning.
In the few-shot setting, decoder-only models achieve at least \(16.39\%\) higher AP than in the zero-shot setting.
\paragraph{Next intent prediction.}
The next intent prediction task yields results similar to those seen in the intent classification task.
As in the intent classification task, decoder-only models cannot match the performance of the fine-tuned encoder models, with a difference of at least \(29.78\%\) AP.
In this task, hierarchical fine-tuning degrades the performance of the encoder models in both metrics by a large margin.
We observe, that AP for GatorTronS drops by \(114.89\%\).
Phi-4 and Qwen2.5 exhibit identical performance, with Llama3.1 trailing behind.
Notably, the decoder-only models do not benefit as much from additional examples as in the prior task.
For Phi-4, the AP difference between zero-shot and few-shot is only \(13.33\%\).
The AUROC performance of Qwen2.5 in the few-shot setting is even lower than in the zero-shot setting, suggesting that intent trajectories can vary significantly.
Providing similar examples may cause confusion rather than offering meaningful support.

\subsection{Intent Performance Analysis}
\label{subsec:intent_perf}

\begin{figure*}[h!]
    \centering
    \includegraphics[width=\linewidth]{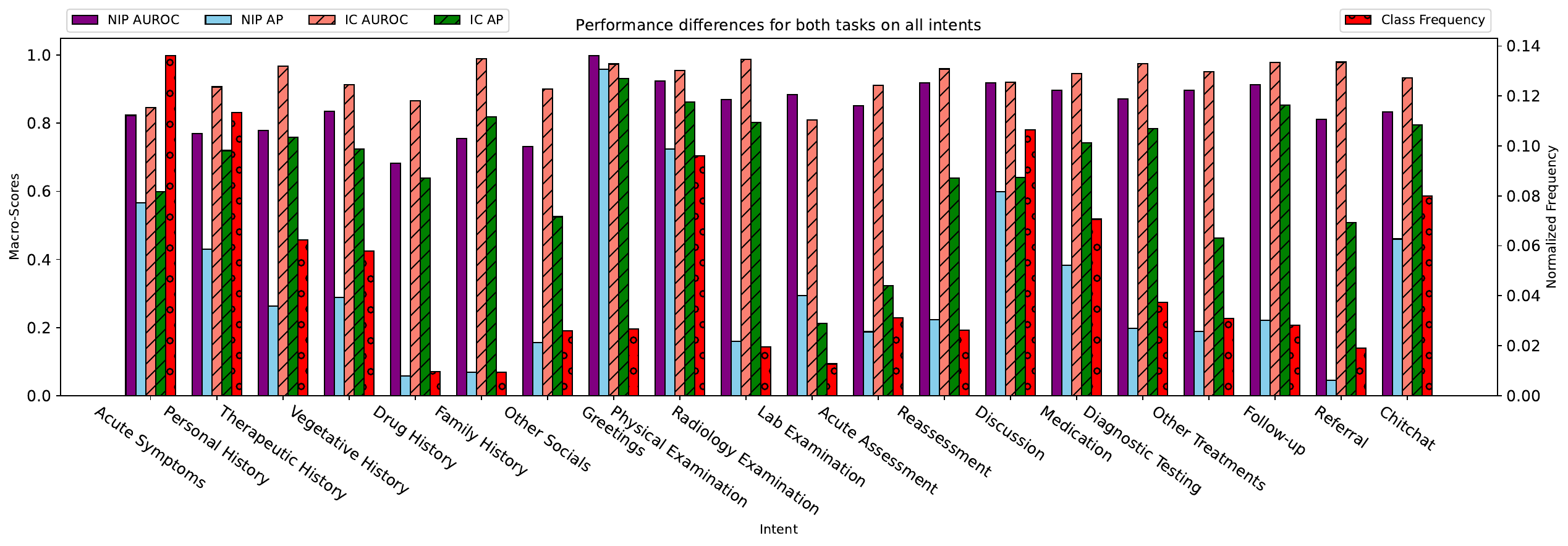}
    \caption{GatorTronS AUROC and AP performance across all intents for both tasks, organized by SOAP categories. The next intent prediction task exhibits a stronger correlation with the present imbalance, whereas no such trend is observed in intent classification.}
    \label{fig:nip_ic_score_plot}
\end{figure*}

\paragraph{Tasks differences in robustness towards intent imbalance.} 
Table \ref{table:benchmark_results} highlights a discrepancy between AUROC and AP for all models in both tasks. 
Although the model performs well on average, classification accuracy decreases across the different intents, indicating reduced performance for less frequent or more challenging intent categories. 
Figure \ref{fig:nip_ic_score_plot} shows the AUROC and AP scores for both tasks per intent, as well as their frequency in the data. 
In the figure, we order the intents according to the SOAP categories, starting with \textit{Subjective} intents on the left, moving through \textit{Objective} and \textit{Assessment}, and ending with \textit{Plan} intents. 
\textit{Chitchat} intents are placed on the far right. 

Our results demonstrate that the frequency of intents has a negligible effect on the AP for the intent classification task.
However, we observe a clear correlation between AP and intent frequency in the next intent prediction task.
This indicates that intent classification is more resilient to intent imbalance, while next intent prediction is significantly affected by this imbalance. 
We explain this divergent behavior with the complexity of task input.
In intent classification, the model only classifies a single utterance, making it less sensitive to intent imbalance.
In contrast, next intent prediction requires the model to understand a trajectory of doctor-patient turns, where intent sequences can vary.
This variability means the model needs more examples to effectively capture the potential intent combinations, making it more susceptible to class imbalance.

\paragraph{Semantic similarities impact intent classification.}
The \textit{Lab Examination} intent has the lowest AP (\(0.21\)) in the intent classification task.
In contrast, the other two \textit{Objective} intents, \textit{Physical Examination} and \textit{Radiology Examination}, achieve significantly higher AP scores of \(0.86\) and \(0.80\), respectively. 
The \textit{Lab Examination} intent and \textit{Radiology Examination} intent share similar semantic structures, since both involve the examination of diagnostic tests.
A closer look into the results reveals that the model misclassifies \textit{Lab Examination} instances as \textit{Radiology Examination} and \textit{Physical Examination}.
This indicates that the model learns to identify the presence of diagnostic tests, but does not distinguish the subtle differences between certain types of tests.
However, in the next intent prediction task, we do not observe such behavior.
The different behavior signifies that the two tasks learn to represent the same intents differently.
Thus, each task poses distinct challenges, even though they share the same intent classes.

\subsection{Error Analysis and Reconstructing Dialogue Sequences}
\label{subsec:diareco}
\begin{figure*}[!ht]
    \centering
    \includegraphics[width=\textwidth]{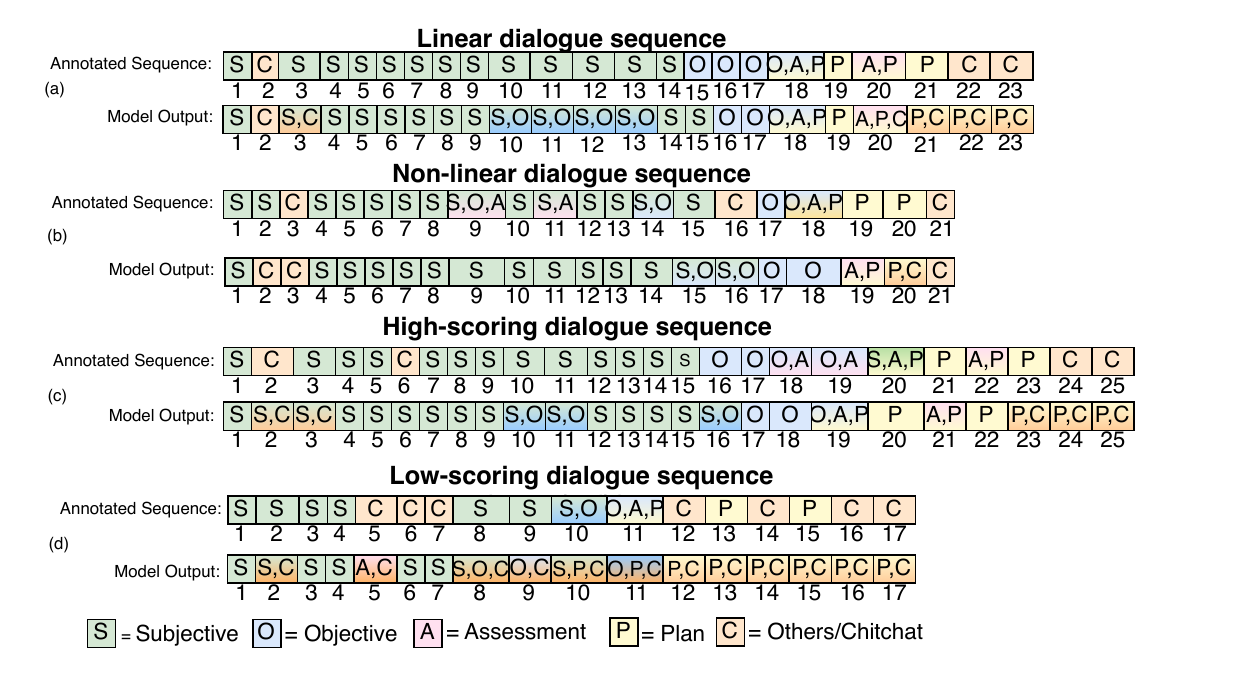}
    \caption{Comparison between a linear, non-linear, high-scoring sequence, and low low-scoring sequence.
    In all cases, the model can replicate the sequence to some extent but fails to reconstruct non-linear sequences. We identify model confidence and phase transitions as major challenges.}
    \vspace{1em}
    \label{fig:highvlow}
\end{figure*}

We do not present the models with a complete dialogue during the next intent prediction training.
However, the ability to comprehend and plan dialogues is essential for models designed to support doctors throughout the differential diagnosis process.
To investigate whether a model fine-tuned on next intent prediction retains these characteristics, we evaluate its ability to reconstruct dialogue sequences.
\paragraph{Dialogue type impacts reconstruction accuracy.}
The cause of a patient visiting a doctor determines the type of interview they conduct.
We categorized the interviews into two types: linear and non-linear.
A dialogue structure is considered linear when the patient presents a common complaint that follows standard examination patterns.
These patterns are characterized by distinct transitions across the SOAP phases as detailed in Section \ref{subsec:phases}.
Cases such as \textit{follow-ups} or \textit{annual exams} dialogues are non-linear,  as the transitions through the SOAP phases do not follow standard patterns.
We show examples of sequences for both types of dialogue in Figure \ref{fig:highvlow}.
In the linear dialogue, we observe that after 14 turns in the symptom-taking phase (\textit{Subjective)}, the physician transitions to the examination phase (\textit{Objective)}, followed by the clinical assessment phase (\textit{Assessment)} and several turns dedicated to the treatment-planning phase (\textit{Plan)}.
The conversation ends with some \textit{Chitchat}.
As for the non-linear dialogue, we do not have distinct transitions between the SOAP phases. 
In turns 9 and 11, the doctor initiates a clinical assessment phase that does not lead to the treatment-planning phase, but to a symptom-taking phase.
We observe the same for the examination phase.
The doctor examines the patient in between the symptom-taking phase instead of conducting the examinations in a coherent sequence of turns.

In summary, the model can reconstruct the sequence of linear dialogues.
However, the model fails to predict anomalies for the non-linear dialogue.
Instead, it defaults to predicting a linear trajectory.

\label{subsec:dialogues}

\paragraph{Model overconfidence limits precision.}
We show additional examples of reconstructed sequences in Figure \ref{fig:highvlow}.
Specifically, we show a comparison between a high-scoring sequence and a low-scoring sequence in terms of average precision.
In both examples, we see that the model predicts more intents than are actually annotated in the data.
Furthermore, the model has a tendency to continue sequences as \textit{Chitchat}.

\paragraph{Model does not learn phase transitions.}
In all the examples shown in Figure \ref{fig:highvlow}, the model does not predict the phase transitions in the correct turns.
We identify two transition error classes.
The first one is that the model predicts a transition one turn too late.
This indicates that the model depends on the content of the previous turns to change phases rather than on learned trajectories.
The second error class is a premature transition by the model, especially present in phase transitions from \textit{Subjective} to \textit{Objective}.
Despite the fact that the doctor has not concluded the symptom-taking phase, the model wants to transition to the objective examination phase after 9 turns.

%% file: sections/05_diag_summarization.tex
\section{Impact of Intent Classification on Medical Dialogue Summarization}
To evaluate the effectiveness of models trained on our intent classification dataset, we integrate them into downstream summarization tasks as outlined in \citet{aci}. 
We test on all five summarization tasks: \textit{full note}, \textit{subjective}, \textit{objective exam}, \textit{objective results}, and \textit{assessment and plan}. 
Each task takes a doctor-patient dialogue as input and generates a medical note.
For instance, in the \textit{subjective} task, we only summarize the subjective findings of the patient, whereas in the \textit{assessment and plan} task, we summarize the diagnosis and treatment plans.

\paragraph{Proposed methodology.}
A fine-tuned intent classification model filters the input dialogue before summarization; we use the best performing GatorTronS from Section \ref{sec:resultsanddiss}. 
The filter removes non-medical utterances or retains those relevant to the specific note categories.
For full-note summarization, we discard utterances classified as \textit{Chitchat} and retain all others. 
In subjective summarization, only \textit{Subjective} utterances are kept.
Objective exam summarization includes \textit{Objective} category utterances with a \textit{Physical Examination} intent. 
Objective results summarization also retains \textit{Objective} category utterances but requires the \textit{Lab Examination} and/or the \textit{Radiology Examination} intent. 
Finally, the assessment and plan summarization keeps only utterances from the \textit{Assessment} or \textit{Plan} categories.

\paragraph{Experimental setup.}
We fine-tune a BART-large \cite{bart} using the hyperparameters from \citet{aci} and employ the same decoder-only models as in the intent classification experiments, with the addition of GPT-4o. 
For decoder-only models, we set the temperature to 0.2 and limit new tokens to 512 for full-note summarization and 256 for section-level summarization. 
We infer them in a zero-shot and few-shot setting, with BM25 as the candidate retriever and 3 candidates per sample.
Additional candidates consist of dialogue and summary.
Performance is reported using F1-macro for Rouge-1, Rouge-2, Rouge-Lsum \cite{rouge}, Medcon \cite{quickumls}, BERTScore \cite{bertscore}, and the average for all metrics.

\paragraph{Experimental results.}
We report in Table \ref{table:summarization-task} results only for the BART model and the best-performing model per task. We provide the full result tables in the supplementary materials.
Decoder-only models in the few-shot setting consistently achieve the highest scores, outperforming the zero-shot setting averaged across all tasks by \(28.93\%\) and the fine-tuned BART by \(63.17\%\).
The significant performance gap between the decoder-only models and BART is twofold.
First, the training data consists of too few samples and too much variance; consequently, the training signal is too coarse for effective fine-tuning.
Second, the average dialogue length in Aci-bench exceeds the 1024 maximum input length of BART; thus, the model has to truncate the input and omit information.
Filtering generally improves the performance of decoder-only models by \(5.39\%\). 
The filter significantly decreases the performance for BART in \textit{full-note} and \textit{subjective} summarization by \(21.05\%\) and \(50\%\), respectively, but improves the decoder-only models in those tasks by \(1.63\%\) and \(10\%\). 
The largest improvement occurs in \textit{objective exam} summarization with an increase of \(72.22\%\) for BART and \(15.38\%\) for the decoder-only model.

\begin{table}[!ht]
\caption{Rouge-* (R-*), Medcon (MC), BERTscore (BS), Average (AVG). Results for the BART model and the best-performing model on the summarization tasks. The scores of the decoder-only models are in the few-shot (3) setting. BART scores on average lowest on all tasks. GPT-4o is not always the best-performing model. The benefit of the filtering is ambivalent for the different model types on the different tasks. We observe the most gains for the \textit{subjective} and \textit{objective exam} tasks.}
\scalebox{1.0}{
\begin{tabular}{cclcccl}
\hline
\textbf{Model}  & \textbf{R-1}  & \textbf{R-2}  & \textbf{R-L}  & \textbf{MC}   & \textbf{BS}   & \textbf{AVG}  \\ \hline
                & \multicolumn{6}{c}{\textbf{Full-Note}}                                                        \\
BART            & \textit{0.37} & \textit{0.14} & \textit{0.14} & \textit{0.42} & \textit{0.84} & \textit{0.38} \\
BART+Filter     & 0.32          & 0.35          & 0.10          & 0.10          & 0.83          & 0.30          \\
Phi-4           & \textbf{0.60} & \textbf{0.60} & 0.55          & 0.65          & \textbf{0.90} & 0.60          \\
Phi-4+Filter    & \textbf{0.60} & \textbf{0.60} & \textbf{0.56} & \textbf{0.68} & \textbf{0.90} & \textbf{0.62} \\ \hline
                & \multicolumn{6}{c}{\textbf{Subjective}}                                                       \\
BART            & \textit{0.39} & \textit{0.20} & \textit{0.32} & \textit{0.45} & \textit{0.87} & \textit{0.46} \\
BART+Filter     & 0.19          & 0.00          & 0.17          & 0.05          & 0.76          & 0.23          \\
GPT-4o          & 0.47          & 0.21          & 0.41          & 0.55          & 0.88          & 0.50          \\
GPT-4o+Filter   & \textbf{0.51} & \textbf{0.25} & \textbf{0.45} & \textbf{0.62} & \textbf{0.90} & \textbf{0.55} \\ \hline
                & \multicolumn{6}{c}{\textbf{Objective Exam}}                                                   \\
BART            & 0.09          & 0.00          & 0.07          & 0.00          & 0.87          & 0.18          \\
BART+Filter     & \textit{0.26} & \textit{0.10} & \textit{0.24} & \textit{0.10} & \textit{0.85} & \textit{0.31} \\
Phi-4           & 0.49          & 0.28          & 0.45          & 0.52          & 0.87          & 0.52          \\
Phi-4+Filter    & \textbf{0.53} & \textbf{0.39} & \textbf{0.56} & \textbf{0.59} & \textbf{0.91} & \textbf{0.60} \\ \hline
                & \multicolumn{6}{c}{\textbf{Objective Results}}                                                \\
BART            & 0.19          & 0.03          & 0.19          & 0.0           & 0.81          & 0.24          \\
BART+Filter     & \textit{0.26} & \textit{0.13} & \textit{0.25} & \textit{0.17} & \textit{0.88} & \textit{0.33} \\
Llama3.1        & 0.26          & \textbf{0.15} & 0.25 & \textbf{0.24} & \textbf{0.85} & \textbf{0.35} \\
Llama3.1+Filter & \textbf{0.29} & 0.12          & \textbf{0.27} & 0.19 & \textbf{0.85} & 0.34          \\ \hline
                & \multicolumn{6}{c}{\textbf{Assessment and Plan}}                                              \\
BART            & 0.35          & 0.10          & 0.28          & 0.18          & 0.85          & 0.35          \\
BART+Filter     & \textit{0.39} & \textit{0.15} & \textit{0.29} & \textit{0.31} & \textit{0.86} & \textit{0.40} \\
GPT-4o          & \textbf{0.48} & 0.21          & \textbf{0.43} & \textbf{0.52} & 0.88          & 0.50          \\
GPT-4o+Filter   & \textbf{0.48} & \textbf{0.22} & \textbf{0.43} & \textbf{0.52} & \textbf{0.89} & \textbf{0.51} \\ \hline
\end{tabular}

}
\centering

\label{table:summarization-task}
\end{table}

\paragraph{Experimental results.}

\paragraph{Qualitative assessment of filter effectiveness.} 
To assess the effectiveness of intent filtering for summary generation, we perform a comparison between all GPT-4o outputs and the reference summaries.
We chose GPT-4o for this evaluation, as it produces the most consistent results across all summarization tasks.
We provide examples in the supplementary material.
\begin{itemize}
    \item \textbf{Reduction of verbosity in summaries.}  
    Since we exclude unwanted information in the dialogue and reduce noise in the input, the filter reduces the verbosity of the generated notes in all summarization tasks.
    We observe the greatest impact on the \textit{objective exam} task. 
    In this task, we summarize the \textit{Physical Examination} (PE) findings and notes are usually very short.
    In addition, we see improvements in \textit{full-note} summarization for chitchat-heavy dialogues.
    \item \textbf{Utterance complexity determines filtered dialogue density.} 
    The intent classification characteristics described in Section \ref{subsec:intent_perf} also apply to the summarization tasks.
    The filter achieves high coverage for utterances in the \textit{subjective} phase, thus it is able to create dense input dialogues and increase summarization quality. 
    The same applies to PE utterances in the \textit{objective exam} summarization, where we observe significant improvements.
    Improvement in \textit{assessment and plan} summarization is only marginal, because the corresponding utterances are long and with overlapping intents.
    The filter does not dissect these utterances for the important information. 
    Noise persists in the input dialogue and reduces the potential summarization quality.
    \item \textbf{Information loss due to incorrect classification.} 
    The filtering model occasionally misclassifies utterances, leading to the omission of relevant information. 
    In such cases, the filtered input dialogues are incomplete, and the summaries perform worse than their unfiltered counterparts.
    The \textit{objective results} summarization highlights this behavior.
    This category focuses on \textit{Radiology-} and \textit{Lab Examination} (LE) utterances.
    As discussed in Section \ref{subsec:intent_perf}, the model has a tendency to misclassify LE utterances as \textit{Physical Examination} (PE).
    Since we filter PE utterances for this category, we lose valuable information and score worse than the unfiltered summarization.
\end{itemize}

In summary, filtering improves performance, particularly for SOAP category-specific summarization, by creating dense input dialogues, which reduces the verbosity in the summary.
We see that this works well for categories in which utterances are less complex, but not as well for categories with more complex utterances.
However, incorrect classification can significantly degrade performance if key utterances are excluded from the input dialogue.

%% file: sections/07_closing.tex
\section{Conclusion}
\label{sec:future}
In this work, we present "Where does it hurt?" - a novel 
medical intent classification dataset for dialogues.
We introduce the complete annotation process and describe the taxonomy based on the SOAP framework. 
This adaptation of the SOAP framework for dialogues allows us to conclude that physicians spend the most turns on subjective symptom-taking, but talk the most during treatment-planning.
Furthermore, we conduct extensive experimental studies on an intent classification task and a next intent prediction task.
We show that classically fine-tuned encoder-only models perform best in both tasks. 
Language models learn to classify doctor utterances to medical intents but struggle to predict the next intent for a sequence of doctor-patient turns.
We examine the robustness of medical intent classification models towards class imbalance and present challenges in reconstructing dialogue trajectories with next intent prediction models.
Lastly, we utilize a model trained on our dataset as a filter in a downstream summarization task and show improved summarization performance against baselines.

\paragraph{Future Work.} First, the dialogue reconstruction experiment in Section \ref{subsec:diareco} shows that the models learn to follow trajectories but fail to identify category transitions. 
Further investigation to improve transition capabilities can lead to better overall reconstruction quality. 
Second, the findings that we acquire on physician behavior during dialogues and common intent trajectories can be utilized to create more sophisticated dialogue generation methods, 
especially in the context of medical note-to-dialogue transcription.

\paragraph{Limitations.} First, we source the dialogues for the annotation from the popular Aci-bench benchmark dataset~\cite{aci}, where the dialogues are role-played and thus do not need further de-identification, and as such may not properly reflect a real-world scenario. 
Second, we fine-tune the encoder models in our experiments, but do not fine-tune the decoder-only models because of computational and budget constraints. Therefore, the performance comparison may unfairly favor the encoder models. 